%% file: tdp2020.tex
\documentclass[runningheads,a4paper]{llncs}
\usepackage{amssymb}
\usepackage[pdftex]{graphicx}
\usepackage{amssymb}
\usepackage[utf8]{inputenc}
\usepackage{url}
\usepackage{float}
\usepackage{amsmath}
\usepackage{graphicx}
\usepackage{wrapfig}
\usepackage{booktabs}
\usepackage{multirow}
\usepackage{lipsum}

\makeatletter
\def\Hline{%
\noalign{\ifnum0=`}\fi\hrule \@height 1pt \futurelet
\reserved@a\@xhline}
\makeatother

\begin{document}

\title{Hibikino-Musashi@Home\\2020 Team Description Paper}

\author{
Tomohiro Ono
\and Yuichiro Tanaka
\and Yutaro Ishida
\and Yushi Abe
\and Kazuki Kanamaru
\and Daichi Kamimura
\and Kentaro Nakamura
\and Yuta Nishimura
\and Shoshi Tokuno
\and Yuya Mii
\and Morio Yamauchi
\and Yuichiro Uemura
\and Takunori Hashimoto
\and Yugo Nakamura
\and Issei Uchino
\and Daiju Kanaoka
\and Takeru Hanyu
\and Kenta Tsukamoto
\and Takashi Morie
\and Hakaru Tamukoh
}
\authorrunning{Tomohiro Ono et al.}
\institute{
Graduate school of life science and systems engineering,\\
Kyushu Institute of Technology,\\
2-4 Hibikino, Wakamatsu-ku, Kitakyushu 808-0196, Japan,\\
\email{hma@brain.kyutech.ac.jp} \\
\url{http://www.brain.kyutech.ac.jp/~hma/wordpress/}
}

\maketitle


\begin{abstract}
Our team, Hibikino-Musashi@Home (HMA), was founded in 2010.
It is based in Japan in the Kitakyushu Science and Research Park.
Since 2010, we have annually participated in the RoboCup@Home Japan Open competition in the open platform league (OPL).
We participated as an open platform league team in the  2017 Nagoya RoboCup competition and as a domestic standard platform league (DSPL) team in the 2017 Nagoya, 2018 Montreal, and 2019 Sydney RoboCup competitions.
We also participated in the World Robot Challenge (WRC) 2018 in the service-robotics category of the partner-robot challenge (real space) and won first place.
Currently, we have 20 members from eight different laboratories within the Kyushu Institute of Technology.
In this paper, we introduce the activities that have been performed by our team and the technologies that we use.
\end{abstract}


\section{Introduction}
Our team, Hibikino-Musashi@Home (HMA), was founded in 2010, and we have been competing annually in the RoboCup@Home Japan Open competition in the open platform league (OPL).
Our team is developing a home-service robot, and we intend to demonstrate our robot in this event in 2020 to present the outcome of our latest research.

In RoboCup 2017 Nagoya, we participated both in the OPL and the domestic standard platform league (DSPL) and in the RoboCup 2018 Montreal and RoboCup 2019 Sydney we participated in the DSPL.
Additionally, in the World Robot Challenge (WRC) 2018, we participated in the service-robotics category of the partner-robot challenge (real space).

In the RoboCup 2017, 2018 and 2019 competitions and in the WRC 2018, we used a Toyota human support robot (HSR) \cite{toyota_hsr}. We were awarded the first prize at the WRC 2018 and third prize at the RoboCup 2019.

In this paper, we describe the technologies used in our robot.
In particular, this paper outlines our object recognition system that uses deep learning \cite{hinton2006fast}, improves the speed of HSR, and has a brain-inspired artificial intelligence model, which was originally proposed by us and is installed in our HSR.


\section{System overview}
Figure \ref{fig:softOverview} presents an overview of our HSR system.
We have used an HSR since 2016.
In this section, we will introduce the specifications of our HSR.

\subsection{Hardware overview}
We participated in RoboCup 2018 Montreal and 2019 Sydney with this HSR.
The computational resources built into the HSR were inadequate to support our intelligent systems and were unable to extract the maximum performance from the system.
To overcome this limitation, using an Official Standard Laptop for DSPL that can fulfill the computational requirements of our intelligent systems has been permitted since RoboCup 2018 Montreal.
We use an ALIENWARE (Intel Core i7-8700K CPU, 32GB RAM, and GTX-1080 GPU) as the Official Standard Laptop for DSPL.
Consequently, the computer equipped inside the HSR can be used to run basic HSR software, such as sensor drivers, motion planning and, actuator drivers.
This has increased the operational stability of the HSR.
%

\subsection{Software overview}
\begin{figure}[bt]
\begin{center}
\includegraphics[scale=0.5]{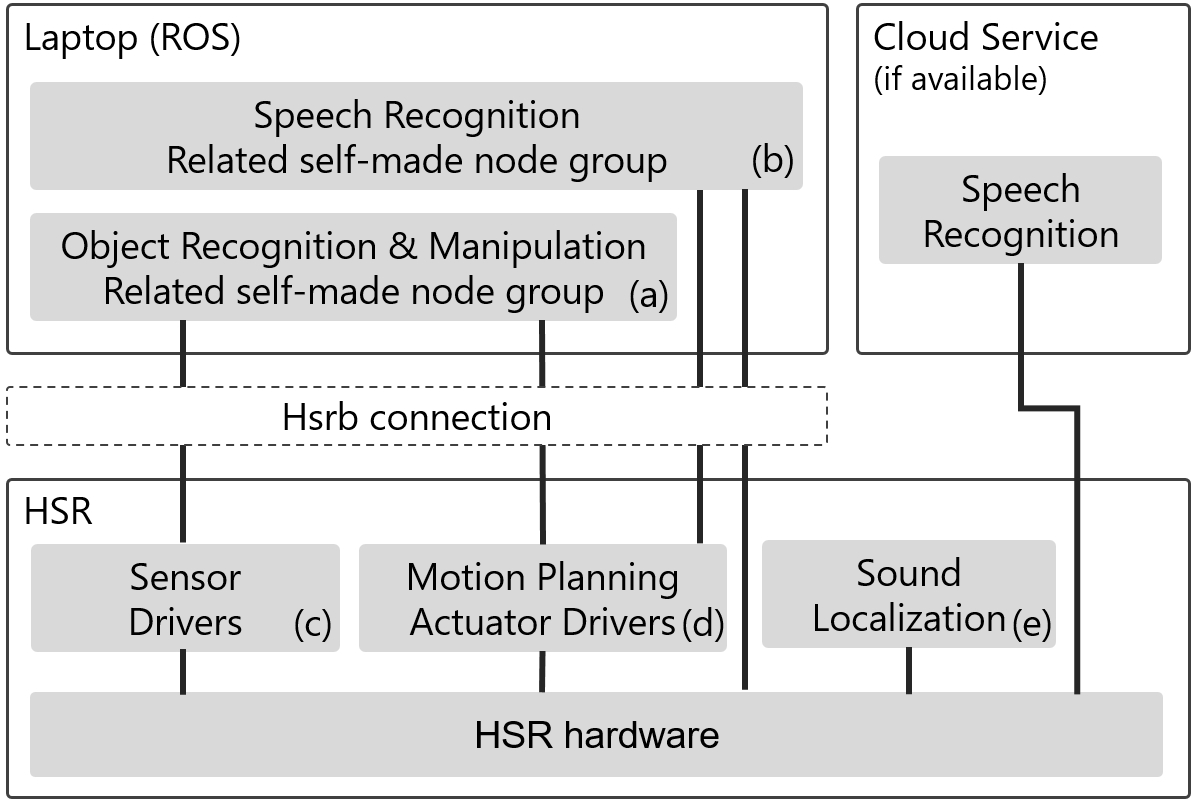}
\caption{Block diagram overview of our HSR system. [HSR, human-support robot; ROS, robot operating system]}
\label{fig:softOverview}
\end{center}
\end{figure}
In this section, we introduce the software installed in our HSR.
Figure \ref{fig:softOverview} shows the system installed in our HSR.
The system is based on the Robot Operating System \cite{ros}.
In our HSR system, laptop computer and a cloud service, if a network connection is available, are used for system processing.
The laptop is connected to a computer through an Hsrb interface.
The built-in computer specializes in low-layer systems, such as HSR sensor drivers, motion planning, and actuator drivers, as shown in Fig. \ref{fig:softOverview} (c) and (d).
Furthermore, the built-in computer has a sound localization system that use HARK \cite{hark}, as shown in Fig. \ref{fig:softOverview} (e).


\section{Object recognition}
In this section, we explain the object recognition system (shown in Fig. \ref{fig:softOverview} (a)), which is based on you look only once (YOLO) \cite{redmon2016you}.

To train YOLO, a complex annotation phase is required for annotating labels and bounding boxes of objects.
In the RoboCup@Home competition, predefined objects are typically announced during the setup days right before the start of the competition days.
Thus, we have limited time to train YOLO during the competition, and the annotation phase impedes the use of the trained YOLO during the competition days.

\begin{figure}[b]
\begin{center}
\includegraphics[scale=0.5]{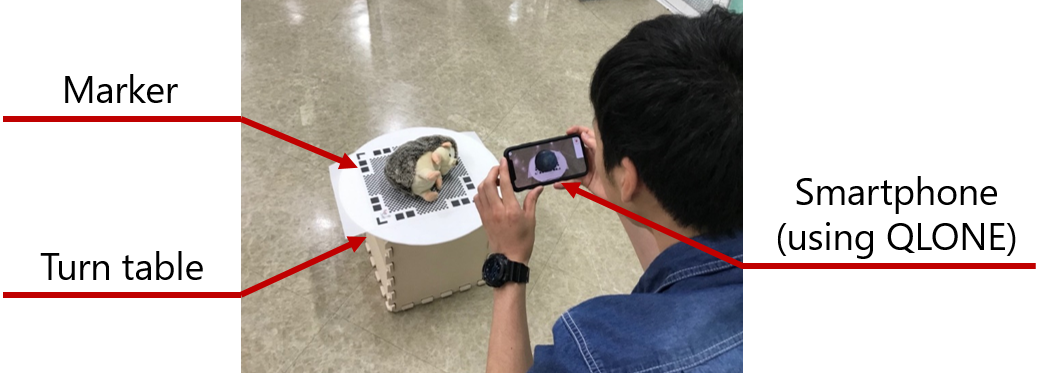}
\caption{Overview of proposed autonomous annotation system for YOLO.}
\label{fig:annotation1}
\end{center}
\end{figure}

We utilize an autonomous annotation system for YOLO using a three-dimensional (3D) scanner.
Figure \ref{fig:annotation1} shows an overview of the proposed system.
In this system, QLONE \cite{qlone}, a smartphone application capable of 3D scanning, is used.
QLONE makes it easy to create 3D models by placing objects on dedicated markers and shooting them.
We placed the marker and object on a turntable and created a 3D model.
In this method, the bottom surface of the object could not be shoot; thus, two 3D models can be created for each object by acquiring the flipped upside-down object.

\begin{figure}[tb]
\begin{center}
\includegraphics[scale=0.5]{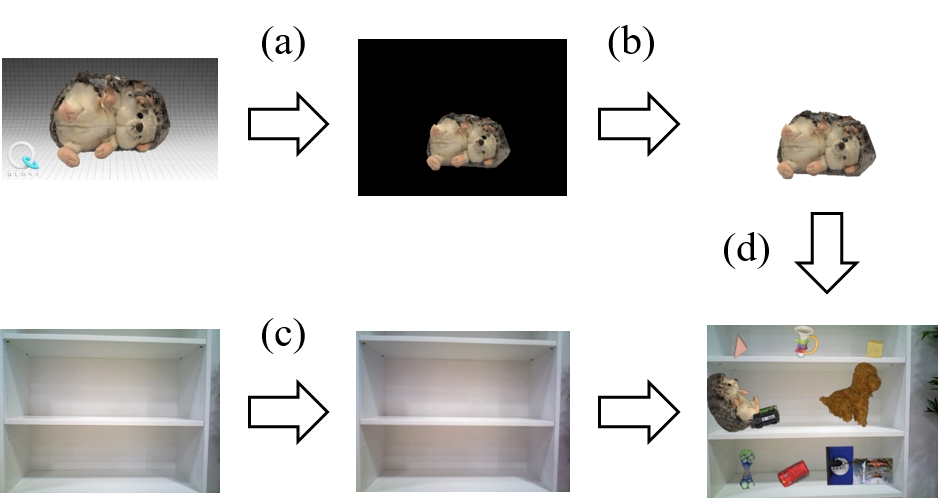}
\caption{Processing flow for generating training images for YOLO.}
\label{fig:annotation2}
\end{center}
\end{figure}

Figure \ref{fig:annotation2} shows the processing flow to generate training images for YOLO.

Multi-viewpoint images are automatically generated from the created two 3D models (Fig. \ref{fig:annotation2} (a)).
Then, we remove image backgrounds (Fig. \ref{fig:annotation2} (b)).

For backgrounds of the training images, we shoot background images, for example, a table, shelf, and other items.
To adapt to various lighting conditions, we apply the automatic color equation algorithm \cite{RIZZI20031663} to the background images (Fig. \ref{fig:annotation2} (c)).
To incorporate the object images into the background images, we define 20-25 object locations on the background images (the number of object locations depends on the background images).
Then, by placing the object images on the defined object locations autonomously, the training images for YOLO are generated (Fig. \ref{fig:annotation2} (d)).
If there are 15 class objects and 306 background images, 400,000 training images are generated.
Additionally, annotation data for the training images are generated autonomously because object labels and positions are known.

Image generation requires ~15 min (using six CPU cores in parallel), and training of YOLO requires approximately 6 h when using the GTX1080 GPU on a standard laptop.
Even though the generated training data are artificial, recognition of YOLO in actual environments works.
The accuracy when learning 10,000 epochs is 60.72\% in a mean average precision (mAP) evaluation.


\section{High-speed behavioral synthesis}
We are working to improve the speed of HSR from two viewpoints: behavioral synthesis and software processing speed.

Regarding behavioral synthesis, we reduce the wasted motion by combining and synthesizing several behaviors for the robot.
For instance, by moving each joint of the arm during navigation, the robot can move to the next action such as grasping without wasting any time as soon as the robot reaches an interim target location. 

Regarding the processing speed, we aim to operate all software at 30 Hz or higher.
To reduce the waiting time for software processing, which causes the robot to stop, the essential functions of the home service robot, such as object recognition and object grasping-point estimation, need to be executed in real time.
We optimized these functions for the Tidy Up Here task. 

\begin{figure}[bt]
  \begin{center}
    \includegraphics[scale=0.65]{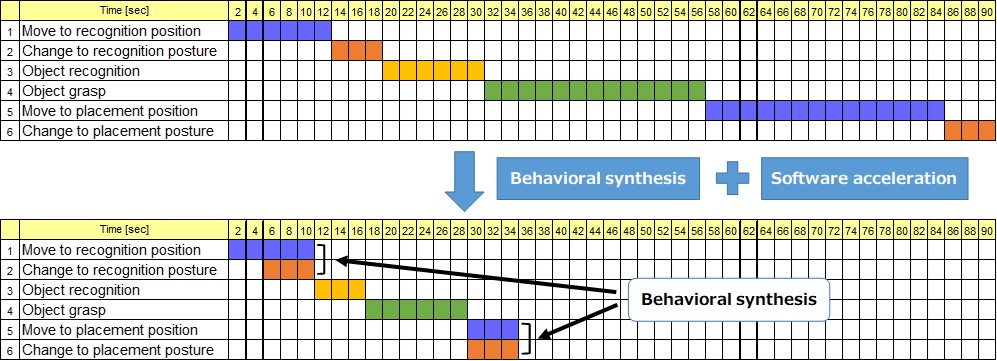}
    \caption{Speed comparison between a conventional system and the proposed high-speed system.}
    \label{fig:synthesis}
  \end{center}
\end{figure}

We used two optimized methods for that task in the WRC 2018 (Fig. \ref{fig:synthesis}). 
In the WRC 2018 results, for which we won first place, our achieved speedup was approximately 2.6 times the prior record. Our robot can tidy up within 34 s per object; thus, so we expect that it can tidy up approximately 20 objects in 15 min. 

\section{Brain-inspired artificial intelligence model}
In this section, we explain a brain-inspired artificial intelligence model that consists of a visual cortex model, an amygdala model, a hippocampus model, and a prefrontal cortex model \cite{tanaka2019biai}.

It is expected that home service robots have local knowledge which is based on the experiences of the robots.
In the case of acquiring local knowledge, its learning is executed during the daily life of the robots.
Thus, applying only deep learning \cite{hinton2006fast} to acquire local knowledge is not effective because the robot cannot prepare big data of the local knowledge.
To acquire local knowledge, we propose an artificial intelligence model that is inspired by the structure of the brain because our brain can acquire local knowledge from only few data.
Mainly, we focus on an amygdala, a hippocampus, and a prefrontal cortex, and the proposed model integrates functions of these parts of the brain.
In addition, we integrated a deep neural network as a visual cortex model into the proposed model.

\begin{figure}[tb]
\begin{center}
\includegraphics[scale=0.4]{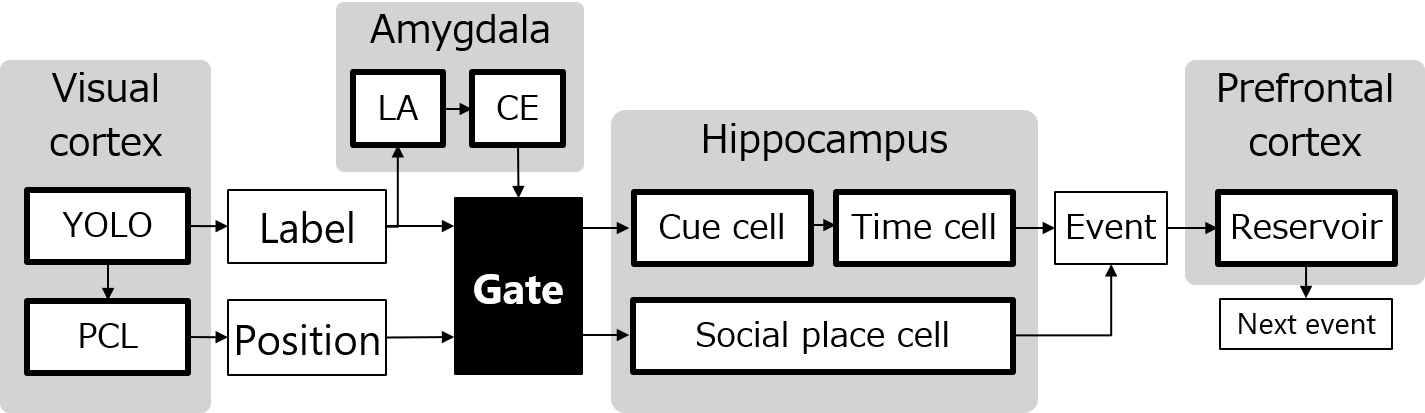}
\caption{A brain-inspired artificial intelligence model that consists of a visual cortex model, an amygdala model, a hippocampus model, and a prefrontal cortex model.}
\label{fig:biai1}
\end{center}
\end{figure}

Figure \ref{fig:biai1} shows the proposed model.
The visual cortex model consists of YOLO and a Point Cloud Library (PCL) \cite{rusu2011pcl}.
The visual cortex model recognizes an environment and outputs the label of detected object and its position.
The object label is input into the amygdala model.
We use the amygdala model for value judgments.
The amygdala model consists of a lateral nucleus (LA) model and a central nucleus (CE) model \cite{tanaka2019amygdala}.
The LA and the CE judge the value of the object.
Only if the value of the object is high enough, the object label and the object position are input into the hippocampus model.
We use the hippocampus model for event coding.
The hippocampus model consists of cue cells, time cells and social place cells as an internal model of detected events.
The cue cells and time cells represent what and when events happen, respectively.
The social place cells represent where events happen.
The cue cells receive the object label and the social place cell receives the object position.
Then, the hippocampus model integrates the outputs of these cells and computes an event vector.
The event vector is input into the prefrontal cortex model.
We use the prefrontal cortex model for event predictions.
The prefrontal cortex model is an echo state network (ESN) \cite{jaeger2001echostate}.
The ESN trains the time-series event vector to predict future events.
After training, the ESN predicts time-series events without input from environments.

\begin{figure}[tb]
\begin{center}
\includegraphics[scale=0.7]{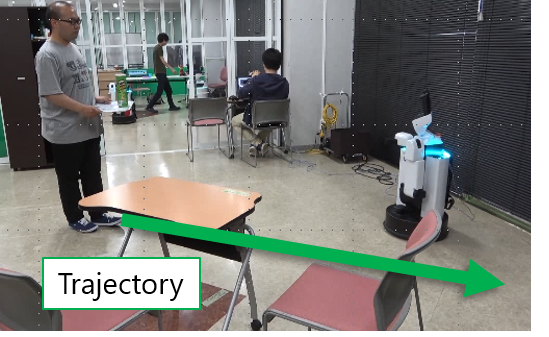}
\caption{Trajectory of a person.}
\label{fig:biai2}
\end{center}
\end{figure}

\begin{figure}[tb]
\begin{center}
\includegraphics[scale=0.7]{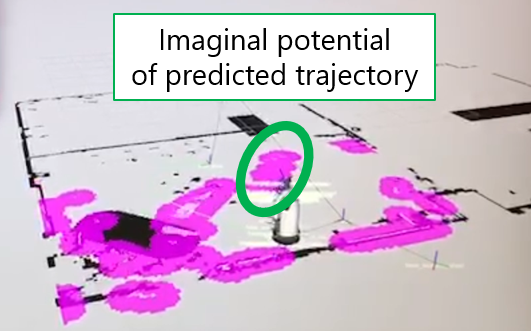}
\caption{Imaginary potential of the predicted trajectory.}
\label{fig:biai3}
\end{center}
\end{figure}

We evaluated the proposed model using the following experiment.
A person walked across in front of a robot, as shown in Fig. \ref{fig:biai2}.
The robot detected the person and learned the trajectory of the person which was a type of local knowledge.
Subsequently, the robot predicted the trajectory.
In addition, the robot added the predicted trajectory as an imaginary potential of a map for SLAM.
Figure \ref{fig:biai3} shows the imaginary potential of the predicted trajectory.
By using the imaginal potential, the robot was able to avoid the person who walked across in front of the robot.
Therefore, the proposed model acquired the local knowledge.

\vspace{-0.3cm}
\section{Conclusions}
In this paper, we summarized the available information about our HSR, which we entered into RoboCup 2019 Sydney.
The object recognition and improved speed of the HSR that we built into the robot were also described.
Currently, we are developing many different pieces of software for an HSR that will be entered into RoboCup 2020 Bordeaux.

\vspace{-0.3cm}
\section*{Acknowledgment}
This work was supported by Ministry of Education, Culture, Sports, Science and Technology, Joint Graduate School Intelligent Car \& Robotics course (2012-2017),
Kitakyushu Foundation for the Advancement of Industry Science and Technology (2013-2015),
Kyushu Institute of Technology 100th anniversary commemoration project : student project (2015, 2018-2019) and YASKAWA electric corporation project (2016-2017),
JSPS KAKENHI grant number 17H01798 and 19J11524,
and the New Energy and Industrial Technology Development Organization (NEDO).

\newpage
\bibliography{tdp2020}
\bibliographystyle{unsrt}

\newpage
\input{RobotDescription}

\end{document}

%% file: RobotDescription.tex



\section*{Appendix 1: Competition results}

\begin{table}[t]
\begin{center}
\caption{Results of recent competitions. [DSPL, domestic standard-platform league; JSAI, Japanese Society for Artificial Intelligence; METI, Ministry of Economy, Trade and Industry (Japan); OPL, open-platform league; RSJ, Robotics Society of Japan]}
\label{tab:result}
\begin{tabular}{l|l|l} \hline
	\multicolumn{1}{c|}{Country} & \multicolumn{1}{c|}{Competition} & \multicolumn{1}{c}{Result} \\ \hline \hline

	Japan & RoboCup 2017 Nagoya & {\bf @Home DSPL 1st} \\
        				      && @Home OPL 5th \\ \hline

        Japan & RoboCup Japan Open 2018 Ogaki & @Home DSPL 2nd \\
        				                      && @Home OPL 1st \\
                                        	         && JSAI Award \\ \hline

        Canada & RoboCup 2018 Montreal & {\bf @Home DSPL 1st} \\
                                         && P\&G Dishwasher Challenge Award \\ \hline

        Japan & World Robot Challenge 2018 & {\bf Service Robotics Category} \\
                                              && {\bf Partner Robot Challenge Real Space 1st} \\
                                         	 && METI Minister's Award, RSJ Special Award \\ \hline

        Australia & RoboCup 2019 Sydney & @Home DSPL 3rd \\ \hline

        Japan & RoboCup Japan Open 2019 Nagaoka & @Home DSPL 1st \\
        				                          && @Home OPL 1st \\ \hline
\end{tabular}
\end{center}
\end{table}

Table \ref{tab:result} shows the results achieved by our team in recent competitions.
We have participated in the RoboCup and World Robot Challenge for several years, and as a result, our team has won prizes and academic awards. \par
Notably, we participated in the RoboCup 2019 Sydney using the system described herein.
We were able to demonstrate the performance of HSR and our technologies.
Thanks to these results, we were awarded the third prize in that competition.

\section*{Appendix 2: Link to Team Video, Team Website}

\begin{itemize}
  \item Team Video \\ \url{https://www.youtube.com/watch?v=0loNuukvOec} \\
  \item Team Website \\ \url{http://www.brain.kyutech.ac.jp/~hma/wordpress/} \\
  \item GitHub \\ \url{https://github.com/hibikino-musashi-athome} \\
  \item Facebook \\ \url{https://www.facebook.com/HibikinoMusashiAthome/} \\
  \item YouTube \\ \url{https://www.youtube.com/channel/UCJEeZZiDXijz6PidLiOtvwQ}
\end{itemize}

\section*{Appendix 3: Robot's Software Description}
For our robot we are using the following software:

\begin{itemize}
	\item OS: Ubuntu 16.04.
	\item Middleware: ROS Kinetic.
	\item State management: SMACH (ROS).
	\item Speech recognition (English):
		\begin{itemize}
			\item rospeex \cite{rospeex}.
			\item Web Speech API.
			\item Kaldi.
		\end{itemize}
	\item Morphological Analysis Dependency Structure Analysis (English): SyntaxNet.
	\item Speech synthesis (English): Web Speech API.
	\item Speech recognition (Japanese): Julius.
	\item Morphological Analysis (Japanese): MeCab.
	\item Dependency structure analysis (Japanese): CaboCha.
	\item Speech synthesis (Japanese): Open JTalk.
	\item Sound location: HARK.
	\item Object detection: point cloud library (PCL) and you only look once (YOLO) \cite{redmon2016you}.
	\item Object recognition: YOLO.
	\item Human detection / tracking:
		\begin{itemize}
			\item Depth image + particle filter.
			\item OpenPose \cite{cao2017realtime}.
		\end{itemize}
	\item Face detection: Convolutional Neural Network.
	\item SLAM: hector\_slam (ROS).
	\item Path planning: move\_base (ROS).
\end{itemize}